\documentclass[letterpaper, 10 pt, conference]{ieeeconf}  

\IEEEoverridecommandlockouts                              

\usepackage{amsmath,amsfonts}
\usepackage{algorithm}
\usepackage{array}
\usepackage[caption=false,font=normalsize,labelfont=sf,textfont=sf]{subfig}
\usepackage{textcomp}
\usepackage{stfloats}
\usepackage{url}
\usepackage{verbatim}
\usepackage{graphicx}
\usepackage{cite}
\usepackage{xcolor}
\usepackage{hyperref}
\usepackage{multirow}
\usepackage{algpseudocode}
\usepackage{tikz}
\usepackage{etoolbox}

\let\labelindent\relax     
\usepackage{enumitem}
\usepackage{circledsteps}

\usepackage{algorithm}
\usepackage{algorithmicx}
\usepackage{algpseudocode}
\usepackage{hyperref}

\title{\LARGE \bf
Synchronized Online Friction Estimation and Adaptive Grasp Control for Robust Gentle Grasp}

\author{Zhenwei Niu$^{1}$, Xiaoyi Chen$^{1}$, Jiayu Hu$^{1}$, Zhaoyang Liu$^{1}$, Jian Tang$^{1}$ and Xiaozu Ju$^{1}$
\thanks{$^{1}$All authors are with X-Humanoid, Beijing, China.
        {\tt\small jensen.niu@x-humanoid.com, ida.chen@x-humanoid.com, selina.hu@x-humanoid.com, leone.liux-humanoid.com, jian.tang@x-humanoid.com and jason.ju@x-humanoid.com}}%
}

\begin{document}

\maketitle
\thispagestyle{empty}
\pagestyle{empty}

\begin{abstract}

We introduce a unified framework for gentle robotic grasping that synergistically couples real-time friction estimation with adaptive grasp control. We propose a new particle filter-based method for real-time estimation of the friction coefficient using vision-based tactile sensors. This estimate is seamlessly integrated into a reactive controller that dynamically modulates grasp force to maintain a stable grip. The two processes operate synchronously in a closed-loop: the controller uses the current best estimate to adjust the force, while new tactile feedback from this action continuously refines the estimation. This creates a highly responsive and robust sensorimotor cycle. The reliability and efficiency of the complete framework are validated through extensive robotic experiments. The website can be found in \href{https://ethan-nzw.github.io/SofeagcGrasp/}{SofeagcGrasp}

\end{abstract}

\section{INTRODUCTION}


Human grasping demonstrates an exceptional ability to achieve both stability, safety and efficiency during object manipulation \cite{johansson2009coding, afzal2022submillimeter}. This proficiency comes from a sophisticated sensorimotor integration process, in which the central nervous system seamlessly combines real-time haptic feedback to dynamically adapt grasp forces \cite{hermsdorfer2011anticipatory}. A pivotal aspect of this is the ability to detect micro-slip events through tactile sensation, enabling the proactive modulation of grip force long before macroscopic sliding occurs \cite{johansson1987signals}. For robots to replicate this "gentle grasp"—applying minimal force sufficient to secure an object without causing damage—they require the ability to perform a quantitative, real-time evaluation of the incipient slip state \cite{westling1984factors, hadjiosif2015flexible, watanabe2007grip, ito2011robust, dongimproved, dong2019maintaining}. 




A promising approach for quantifying the incipient slip state is the use of a "contact factor," a metric derived from micro-element resultant forces that can serve as a direct input for reactive grasp control \cite{sui2023novel, li2025learning}. However, its practical application is impeded by a dependency on a known friction coefficient—a prerequisite that is often infeasible for arbitrary objects. This limitation has led to solutions misaligned with the goal of human-like adaptability. For instance, Li \textit{et al}. \cite{li2025learning} employ a two-stage estimation and control process that is prohibitively slow. While the online estimation method in \cite{sui2023novel} is a conceptual advance, its multi-second convergence time remains significantly slower than human performance and is insufficient for real-time applications.




A fundamental limitation underlying above robotic grasping strategies is the decoupling of friction estimation from grasp control. In contrast, human grasping demonstrates a tightly synchronized process where haptic feedback and motor control are seamlessly integrated to achieve reliable manipulation \cite{johansson2009coding}. To endow robots with similar adaptive capabilities, two primary challenges must be addressed: (1) achieving rapid and robust online friction estimation and (2) developing a control framework that seamlessly integrates this estimation in real time.

Concerning the first challenge, many existing approaches rely on Coulomb’s law with a deterministic friction coefficient $\mu$ \cite{chen2018tactile, cutkosky1986friction, han1996analysis, ma2018friction}. This assumption, however, is inadequate to capture the variability and uncertainty inherent in real-world physical interactions. A more representative approach should model $\mu$ as a probability distribution to better reflect its unpredictable nature. For example, \cite{liu2023beyond} introduces a regression model incorporating a random variable to represent the friction coefficient, thus more effectively accommodating the complexities of practical grasping scenarios. Although this method improves upon stochastic assumptions, it still depends on offline data to calibrate the model for unfamiliar materials. This requirement limits its applicability in unstructured environments and contrasts with the human ability to adapt rapidly and efficiently in interaction.


Regarding the second challenge, although several studies have attempted to integrate friction estimation with grasp control, many still treat these as decoupled processes \cite{li2025learning, sui2023novel, li2025modeling}. For instance, \cite{li2025modeling} proposed an adaptive control framework that employs a regression model similar to \cite{liu2023beyond} for friction estimation, combined with a stick ratio-based grasp controller. However, in this architecture, estimation and control operate sequentially: the system must first complete the friction estimation phase before switching to a controller that uses the inferred parameters. Although this represents a notable advance toward adaptive manipulation, the framework remains fundamentally asynchronous. The need for a separate estimation stage and explicit control-switching mechanism not only increases system complexity but also introduces potential instability during transitions.


\begin{figure*}
    \centering
    \includegraphics[width=0.9\linewidth]{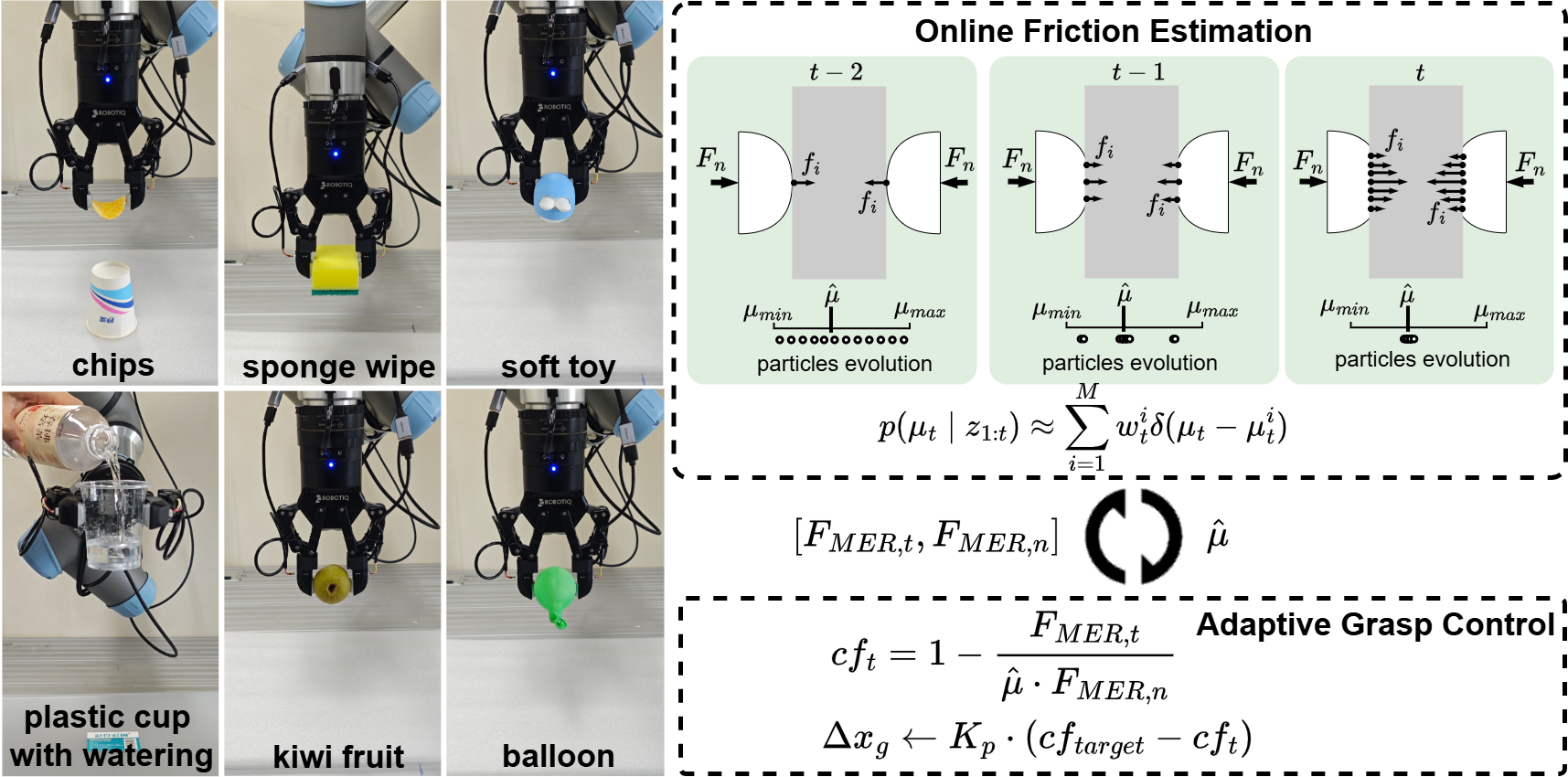}
    \caption{The proposed synchronized architecture, where the particle filter-based estimator is synergistically integrated with a reactive grasp controller, while new tactile feedback from grasp continuously refines the estimation. This allows a close loop for online friction estimation and continuous force modulation, achieving real-time adaptability and grasp stability.}
    \label{fig-whole framework}
\end{figure*}

Therefore, a synchronized strategy is essential to achieve the continuous integration of sensing and control that characterizes human sensorimotor coordination. To this end, we propose a unified framework with the following key contributions:

\begin{itemize}
    \item We propose a new stochastic modeling approach for friction and introduce an online particle filter-based estimation algorithm that enables rapid, robust, and real-time inference of the friction coefficient during robotic manipulation, even under dynamic physical variations.
    \item We develop a tightly coupled estimation-control architecture, where the particle filter-based estimator is synergistically integrated with a reactive grasp controller, while new tactile feedback from grasp continuously refines the estimation. This allows a close loop for online friction estimation and continuous force modulation, achieving real-time adaptability and grasp stability.
    \item We validate the proposed framework through extensive real-world robotic experiments, demonstrating significant performance in both grasp reliability and adaptability.
\end{itemize}

The remainder of this article is structured as follows. Section \ref{sec-grasp model} presents the contact model based on the Vision-Based Tactile Sensor (VBTS). Section \ref{sec-whole method section} details the proposed online friction estimation method and its synergistic integration with adaptive grasp control. Section \ref{sec-experiments} provides comprehensive experimental evaluations and real-time grasping assessments. Finally, Section \ref{sec-conclusion} concludes this article.

\section{Contact Model Based on VBTS}
\label{sec-grasp model}

We use the contact model same as \cite{li2025learning}. For every tactile element at position $p^i = (x^i, y^i, z^i)$ on the vision-based tactile sensor (VBTS) contact surface $\textbf{\textit{S}}(x,y,z)$, the micro-scale force vector is denoted as $\textit{\textbf{f}}^{i}=[f_{x}^{i}, f_{y}^{i}, f_{z}^{i}]$. The unit normal vector $\mathbf{n}^i$ to the surface $\textbf{\textit{S}}$ at point $p^i$ is given by:

\begin{equation}
    \mathbf{n}^i = \left[ n_x^i, n_y^i, n_z^i \right]^T = \frac{1}{\|\nabla S\|} \cdot \left[ \frac{\partial S}{\partial x}, \frac{\partial S}{\partial y}, \frac{\partial S}{\partial z} \right]^T.
\end{equation}

The micro-element normal force $\mathbf{\it{f}}^i_{n}$ and tangential force $\mathbf{\it{f}}^i_{t}$ at the contact point $p^{i}$ can be computed as follows:

\begin{align}
    \mathbf{\it{f}}^i_{n} &= -(\mathbf{\it{f}}^i \cdot \mathbf{\it{n}}^i) \cdot \mathbf{\it{n}}^i \\ 
    \mathbf{\it{f}}^i_{t} &= - (\mathbf{\it{f}}^{i} + (\mathbf{\it{f}}^i \cdot \mathbf{\it{n}}^i) \cdot \mathbf{\it{n}}^i
\end{align}

Based on Coulomb’s friction model, the contact point $p^{i}$ remains in a no-slip state provided that:
\begin{equation}
    \mu > \|\mathbf{\it{f}}^i_{t}\|/\|\mathbf{\it{f}}^i_{n}\|,
\end{equation}
where $\mu$ denotes the coefficient of friction. The above no-slip condition is also applicable to soft objects deforming within the elastic contact assumption.





The contact coefficient is defined as:
\begin{equation}
\label{eq-contact coefficient}
    cf = 1 - \frac{F_{\text{MER},t}}{\mu \cdot F_{\text{MER},n}},
\end{equation}
where $F_{\text{MER},t}$ and $F_{\text{MER},n}$ represent the integral of tangential and normal force magnitudes over the contact surface, respectively. These are mathematical constructs representing resultant forces, not direct physical vectors, and are defined as:

\begin{align}
    F_{\text{MER},n} &= \int_S \|\mathbf{f}^i_n\| \cdot dA = \int_S \|(\mathbf{f}^i \cdot \mathbf{n}^i) \cdot \mathbf{n}^i\| \cdot dA \, , \\
F_{\text{MER},t} &= \int_S \|\mathbf{f}^i_t\| \cdot dA = \int_S \|\mathbf{f}^i - (\mathbf{f}^i \cdot \mathbf{n}^i) \cdot \mathbf{n}^i\| \cdot dA \, .
\end{align}


The contact coefficient, denoted as $cf$, is bounded between 0 and 1. A value of $cf$ close to 0 indicates that the contact state is to macro slip, while values closer to 1 reflect greater contact stability. This metric provides a quantitative measure of grasp security and can be directly incorporated into adaptive grasping control.


\section{Synchronized Online Friction Estimation and Adaptive Grasp Control}
\label{sec-whole method section}
In this section, we first introduce a new online friction estimation method based on a stochastic representation of the friction coefficient. We then tightly couple this estimation process with grasp control to form a unified closed-loop framework, enabling reliable and adaptive robotic grasping under dynamic physical interactions. 

\subsection{Particle Filter for Friction Coefficient Estimation}
\label{sec-particle filter}

We propose an online friction estimation framework based on a particle filter to dynamically identify an object's coefficient of friction during the whole grasp execution. The particle filter is a Bayesian filtering method, which's goal is to track the state of a stochastic system given the sensor observations. 

\begin{algorithm}[t]
\caption{Particle Filter for Friction Coefficient Estimation}
\label{alg:friction_estimation}
\begin{algorithmic}[1]
    \Require $M$: number of particles, $\mu_{\min}$, $\mu_{\max}$: friction bounds, $cf_{\text{target}}$: target contact coefficient
    \Ensure $\hat{\mu}_t$: estimated friction coefficient
    
    \State Initialize particles: $\text{particles} \sim \mathcal{U}(\mu_{\min}, \mu_{\max})$
    \State Initialize weights: $\text{weights} \gets \frac{1}{N}$
    
    \For{$i = 1$ to $M$} \Comment{Prediction step}
        \State $\text{particles}[i] \gets \text{particles}[i] + \mathcal{N}(0, \sigma_p^2)$
        \State $\text{particles}[i] \gets \max(\mu_{\min}, \min(\text{particles}[i], \mu_{\max}))$
    \EndFor
    
    \For{$i = 1$ to $M$} \Comment{Update step}
        \State $\ F_{MER,t}^{\text{expected}} \gets \text{particles}[i] \cdot F_{\text{MER,n}} \cdot (1 - cf_{\text{target}})$
        \State $\text{residual} \gets F_{\text{MER,t}} - F_{MER,t}^{\text{expected}}$
        \State $\text{likelihood} \gets \mathcal{N}(\text{residual}; 0, \sigma^2)$
        \State $\text{weights}[i] \gets \text{weights}[i] \cdot \text{likelihood}$
    \EndFor
    
    \State $\text{weights} \gets \text{weights} / \sum \text{weights}$ \Comment{Normalize weights}
    
    \If{$\frac{1}{\sum (\text{weights}^2)} < \frac{M}{2}$} \Comment{Check effective sample size}
        \State $\text{particles}, \text{weights} \gets \text{Resample}(\text{particles}, \text{weights})$
    \EndIf
    
    \State $\hat{\mu}_t \gets \sum_{i=1}^M \text{weights}[i] \cdot \text{particles}[i]$
\end{algorithmic}
\end{algorithm}

In contrast to treating friction as a fixed, deterministic parameter, we model it as a stochastic variable. Therefore, we define the system state as $\mu_t$. Its state evolution is modeled as a random walk process:

\begin{align}
\label{eq-system model}
    \mu_{t} &= \mu_{t-1} + \epsilon_{t}, \quad \epsilon_{t} \sim \mathcal{N}(0, \sigma_{p}^{2}),\\
    \mu_{t} &= \text{min}(\text{max}(\mu_{t}, \mu_{min}), \mu_{max})
\end{align}
where $\sigma_p^2$ represents process noise variance. Here, $\mu_{min}$ and $\mu_{max}$ are the physical constrains boundary. The state of interest is referred to as the posterior state, $p(\mu_{t} | z_{1:t})$, where $z_{1:t} = {z_1, z_2, \ldots, z_t }$ represents the sequence of all observations from the start of the grasp until time $t$.




While the contact model in (\ref{eq-system model}) generates observations for both normal and tangential forces, we only use tangential force as the observation, $z_{t} = F_{\text{MER},t}$, to simplify the state estimation process and focus on the friction coefficient $\mu_{t}$. The observation model defining the relationship between these elements is:

\begin{equation}
    F_{MER,t}(t) = \mu_{t} \cdot F_{MER,n}(t)(1-cf_{target}) + \nu_{t},
\end{equation}
where $\nu_t \sim \mathcal{N}(0, \sigma_o^2)$ is zero-mean Gaussian observation noise. The parameter $cf_{\text{target}}$ is a predefined contact coefficient. In practice, we set $cf_{\text{target}} = 0.25$ to maintain the force to be 33\% larger than the minimum force required based on the theory in \cite{sui2023novel}.

The estimation begins with a prior distribution, $p(\mu_0)$. Using a Bayesian framework, we recursively update the posterior belief, $p(\mu_{t} | z_{1:t})$. This recursive update is performed in two stages: a prediction step and an update step.

The prediction step is to predict the current distribution $p(\mu_{t} | z_{1:t-1})$ with the previous distribution $p(\mu_{t-1} | z_{1:t-1})$:

\begin{equation}
    p(\mu_{t} | z_{1:t-1}) = \int p(\mu_{t} \mid \mu_{t-1}) p(\mu_{t-1} | z_{1:t-1}) \, d\mu_{t-1},
\end{equation}
where uses the Markov assumption, $p(\mu_{t} \mid \mu_{t-1}, z_{1:t-1})=p(\mu_{t} \mid \mu_{t-1})$. Note that $p(\mu_{t} \mid \mu_{t-1})$ is a system model,  which is illustrated in (\ref{eq-system model}), and describes how the system transitions from previous state to the current state. 

In the second step, the posterior distribution $p(\mu_{t}\mid z_{1:t})$ can be updated based on the new contact force measurement at timestep $t$.

\begin{align}
    p\left(\mu_{t} \mid z_{1:t}\right) &= \frac{p(z_t \mid \mu_{t}, y_{1:t-1})p(\mu_{t} \mid z_{1:t-1})}{p(y_t \mid z_{1:t-1})} \\
&= \frac{p(z_t \mid \mu_{t})p(\mu_{t} \mid z_{1:t-1})}{p(z_t \mid y_{1:t-1})}.
\end{align}


where the contact normalizing factor is:
\begin{equation}
    p(y_t | z_{1:t-1}) = \int p(z_t \mid \mu_{t}) p(\mu_{t} | z_{1:t-1}) \, d\mu_{t}.
\end{equation}

The integration required for this Bayesian update is analytically intractable for nonlinear and non-Gaussian systems. To address this, a Monte Carlo sampling approach is used. The algorithm operates by generating M random samples, called particles, $\chi_{t} = \{ \mu_{t}^{1}, \mu_{t}^{2}, \ldots, \mu_{t}^{M}\}$. Each particle represents a hypothesis for the hidden state (i.e., the friction coefficient). Ideally, if the particles are drawn from the true posterior distribution, $\mu_{t}^{i} \sim p(\mu_{t} \mid z_{1:t})$, the set $\chi_{t}$ constitutes a non-parametric representation of the underlying state distribution. The optimal Bayesian solution is approximated by a weighted sum of these particles:

\begin{equation}
    p(\mu_{0:t} \mid z_{1:t}) \approx \sum_{i=1}^{M}w_{t}^{i} \delta (\mu_{0:t} - \mu_{0:t}^{i}),
\end{equation}
where $\{w_{t}^{i}, \mu_{0:t}^{i}\}_{i=1}^{M}$ is the set of $M$ weighted samples. The weight $w_{t}^{i} \in (0, 1]$ signifies the relative importance of the sample $\mu_{0:t}^{i}$, with $\sum_{i=1}^{M}w_{t}^{i}=1$. Samples with higher weights correspond to more probable estimates of the true state sequence. Here, $\delta(\cdot)$ denotes the Dirac delta function, which is zero everywhere except at the origin and integrates to one. This formulation defines a discrete probability measure that converges to the true posterior distribution.

However, directly sampling from the true posterior $p(\mu_{t} \mid z_{1:t})$ is often infeasible in practice. Instead, particles are drawn from a tractable proposal distribution (or importance density), denoted as $q(\mu_{t} \mid z_{1:t})$. To account for the discrepancy between this proposal distribution and the target posterior, each particle is assigned an importance weight, calculated as:

\begin{equation}
\label{eq-weights calcaulation}
    w_{t}^{i} \propto \frac{p(\mu_{0:t}^{i} \mid z_{1:t})}{q(\mu_{0:t}^{i} \mid z_{1:t})}.
\end{equation}

For the specific application of recursive Bayesian filtering, this general formulation can be simplified under the Markov assumption. The weight update rule is derived recursively to enable efficient computation:

\begin{equation}
\label{eq-weight recursive}
    w^{i}_t \propto w^{i}_{t-1} \frac{p(z_t \mid \mu^{i}_t) p(\mu^{i}_t \mid \mu^{i}_{t-1})}{q(\mu^{i}_t \mid \mu^{i}_{t-1}, z_t)},
\end{equation}
where $w^{i}_{t-1}$ is the previous weight of the $i$-th particle. The posterior distribution at time $t$ is then approximated by the weighted set of particles:

\begin{equation}
    p(\mu_{t} \mid z_{1:t}) \approx \sum_{i=1}^{M}w_{t}^{i} \delta (\mu_{t} - \mu_{t}^{i}),
\end{equation}


The most common choice of the importance density is $p(\mu_{t}^{i} \mid \mu_{t-1}^{i})$, which means:

\begin{equation}
    q(\mu^{i}_t \mid \mu^{i}_{t-1}, z_t) = p(\mu_{t}^{i} \mid \mu_{t-1}^{i}),
\end{equation}
therefore, (\ref{eq-weight recursive}) simplifies to:
\begin{equation}
    w^{i}_t \propto w^{i}_{t-1} p(z_t \mid \mu^{i}_t).
\end{equation}

The likelihood $p(z_t \mid \mu_{t})$ in above process is obtained via:
\begin{equation}
    p(z_{t} \mid \mu_{t}) = \mathcal{N}(z_{t}-\hat{z}_{t}; 0, \sigma_{o}^2),
\end{equation}
where $\hat{z}_{t}$ is the expected micro-element resultant tangential force to be evaluated, ${z}_{t}$ is the true measurement and $\sigma_{o}$ is the variance factor accounting for the measurement uncertainty.

A critical aspect of the particle filter's procedure is monitoring particle degeneracy, a phenomenon where the probability mass becomes concentrated on a small number of particles, while the remaining particles contribute negligibly to the approximation. To mitigate this degeneracy, a resampling step is employed immediately following the weight update. During resampling, a new set of $M$ particles is generated by selecting from the current weighted set with replacement. The selection probability for each particle is proportional to its importance weight. Consequently, particles with higher weights are likely to be duplicated multiple times, while those with low weights are typically discarded. Following resampling, all weights are reset to $1/M$. This process effectively refocuses computational resources on the most promising regions of the state space. Resampling is triggered adaptively based on the degree of degeneracy. The standard metric for quantifying this degeneracy is the Effective Sample Size (ESS), which is approximated by:

\begin{equation}
M_{\text{eff}} \approx \frac{1}{\sum_{i=1}^{M} (w_{t}^{i})^2}.
\end{equation}

To maintain a diverse set of particles, resampling is triggered only when $N_{\text{eff}}$ falls below a predefined threshold (e.g., $M_{\text{eff}} < \frac{M}{2}$). The algorithm for the whole friction coefficient estimation is shown in Algorithm \ref{alg:friction_estimation}.

\subsection{Synchronized Online Friction Estimation and Adaptive Grasp Control}
\label{sec-adaptive grasp}
The process of grasping, lifting, and replacing an object can be decomposed into seven distinct phases: reach, load, lift, hold, replace, unload, and release \cite{johansson2009coding}. 

\begin{algorithm}[t]
\caption{Synchronized Online Friction Estimation and Adaptive Grasp Control}
\label{alg:adaptive grasp}
\begin{algorithmic}[1]
\Require $cf_{\text{target}}$: target contact coefficient, $K_{p}$: proportional gain, $\Delta cf$: contact coefficient tolerance, $F_{n}^{thr}$: contact threshold force

\State Initialize adjustment flag: $\text{contactFlag} \gets FALSE$,

\While{$\mid{F_{n}}\mid < F_{n}^{thr}$}
    \State $x_{g} \gets x_{g} + \Delta d$
    \State $\textit{Gripper.Excuate}(x_{g})$
\EndWhile
\State $\text{contactFlag} \gets TRUE$

\While{contactFlag}
    \State $\hat{\mu} \gets \textit{FricitonEstimation}()$ \Comment{\text{Section}(\ref{sec-particle filter})}
    \State $\hat{cf}_{t} \gets $ \textit{ContactCoefficient}() \Comment{(\text{\ref{eq-contact coefficient}})}
    \If{$\parallel cf_t - cf_{\text{target}}\parallel > \Delta cf$}
        \State $\Delta x_g \gets K_p \cdot (cf_{\text{target}} - cf_t)$
    \Else
        \State $\Delta x_{g} \gets 0$
    \EndIf
    \State $x_{g} \gets x_{g} + \Delta x_{g}$
    \State $\textit{Gripper.Excuate}(x_{g})$
\EndWhile

\end{algorithmic}
\end{algorithm}

During the reach phase, the objective is to establish contact between the object and the tactile sensor-equipped gripper. The gripper closes gradually until a contact is detected, which is determined by a threshold on the resultant normal force: $F_{n} > F_{n}^{\text{thr}}$, where $F_{n}^{\text{thr}}$ is predefined contact force. This threshold is partially generalizable across objects of similar weight but must be adjusted for objects with significant weight differences. For example, a relatively low threshold suffices for lightweight objects such as plastic chips, sponge wipes, and balloons, whereas heavier objects—such as soft toys and kiwifruit—require a higher force threshold to ensure reliable contact detection.

Subsequently, during the load phase, the friction coefficient is estimated using the particle filter method detailed in Section \ref{sec-particle filter}. This estimate then enables the online update of the contact coefficient $cf_{t}$ via (\ref{eq-contact coefficient}), utilizing the measured forces. A proportional control law adjusts the gripper position $x_g$ to drive $cf_t$ towards a predefined target $cf_{\text{target}}$, which defines the desired safety margin against slip. The gripper command is given by:
\begin{equation}
\Delta x_g(t) = K_p \cdot (cf_{\text{target}} - cf_t)
\label{eq-control_law}
\end{equation}
where $K_p$ is a proportional gain. This phase terminates when the contact coefficient is stabilized within a tolerance $\delta$:
\begin{equation}
\label{eq-tolerance}
|cf_t - cf_{\text{target}}| < \delta
\end{equation}

It is noteworthy that the friction estimation module and the adaptive grasp controller form a closed-loop system: the controller utilizes the current best estimate to adjust the gripper position, while new tactile feedback resulting from this adjustment is continuously incorporated to refine the estimation. This reciprocal interaction establishes a highly responsive and disturbance-resistant sensorimotor cycle, effectively mirroring the seamless integration of perception and action observed in human sensorimotor coordination.

\begin{figure*}
    \centering
    \includegraphics[width=\linewidth]{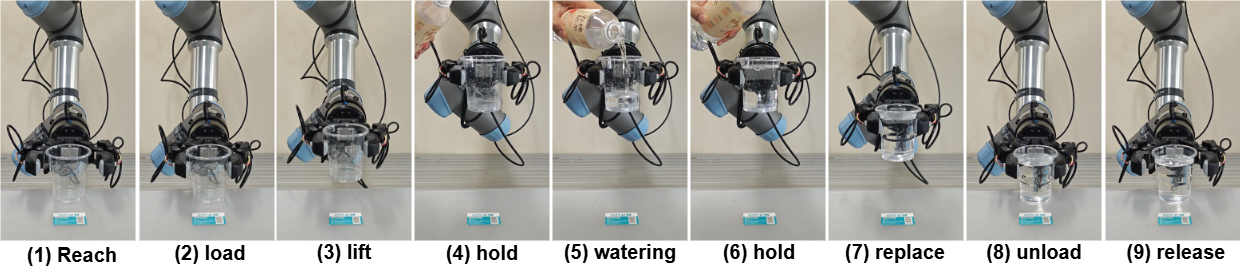}
    \caption{A grasping experiment is conducted using a plastic cup with a moistened surface, while water is incrementally added during the manipulation task.}
    \label{fig-grasp plastic cup}
\end{figure*}

\begin{figure*}
    \centering
    \includegraphics[width=\linewidth]{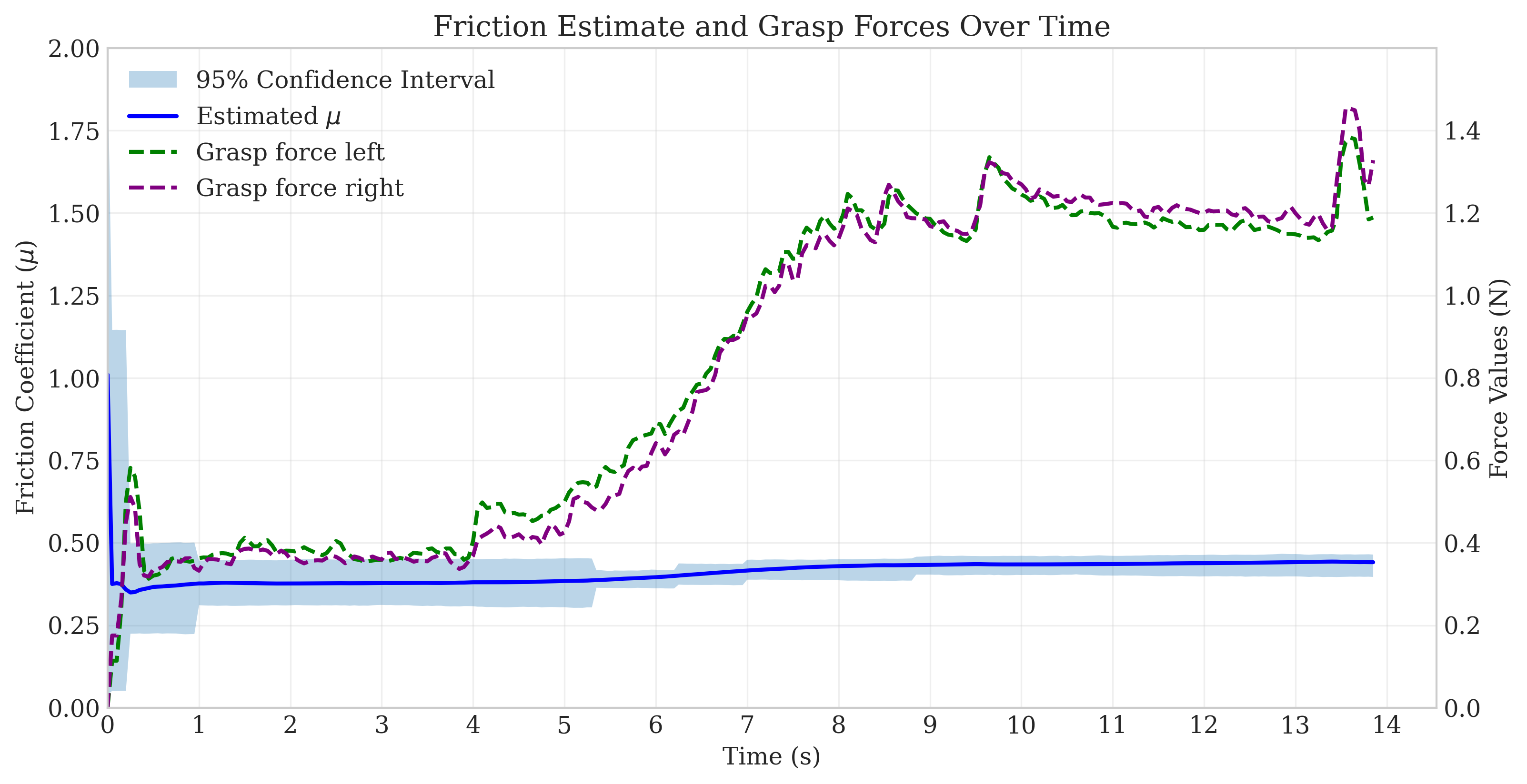}
    \caption{The grasp force and friction estimation results for grasping a plastic cup with unstable friction and changing weight.}
    \label{fig-results plastic cup grasp}
\end{figure*}

During the lift and hold phases, the manipulator executes the motion to elevate the object and maintain it at the desired position. Throughout these critical phases, the stability of the grasp is preserved by continuously applying the control law specified in (\ref{eq-control_law}), governed by the tolerance criterion defined in (\ref{eq-tolerance}). This continuous feedback ensures robust force regulation, preventing slip or excessive deformation despite potential external disturbances or object dynamics. The algorithm for the whole framework of adaptive grasp control is illustrated in Algorithm \ref{alg:adaptive grasp}.

\section{Experiments}
\label{sec-experiments}

In this section, we present a real-world robotic evaluation of the proposed framework for grasping delicate objects. Two distinct experimental scenarios are designed to assess system performance under challenging conditions. First, we evaluate the full seven-phase grasping process but under dynamic variations in object properties, including unstable surface friction and changing weight during manipulation. Second, we further validate the reliability of the system by testing its performance during robot motion involving acceleration and deceleration while securely grasping different delicate objects. 





The experimental platform for evaluating our grasping framework is illustrated in Fig. \ref{fig-grasp plastic cup}. The setup consists of a UR5e robotic arm and a Robotiq gripper integrated with two Vision-Based Tactile Sensors (Tac3D) \cite{zhang2023improving}. The Tac3D sensors deliver high-resolution tactile feedback, including 3D deformation fields, distributed 3D force profiles, and resultant contact forces, enabling real-time estimation of the friction coefficient and closed-loop force adjustment. All processing is performed on a workstation with an Intel i9-13900K CPU (5.80 GHz) and an NVIDIA RTX 3090 GPU (24 GB). The control loop operates at 30 Hz.

\begin{figure*}
    \centering
    \includegraphics[width=\linewidth]{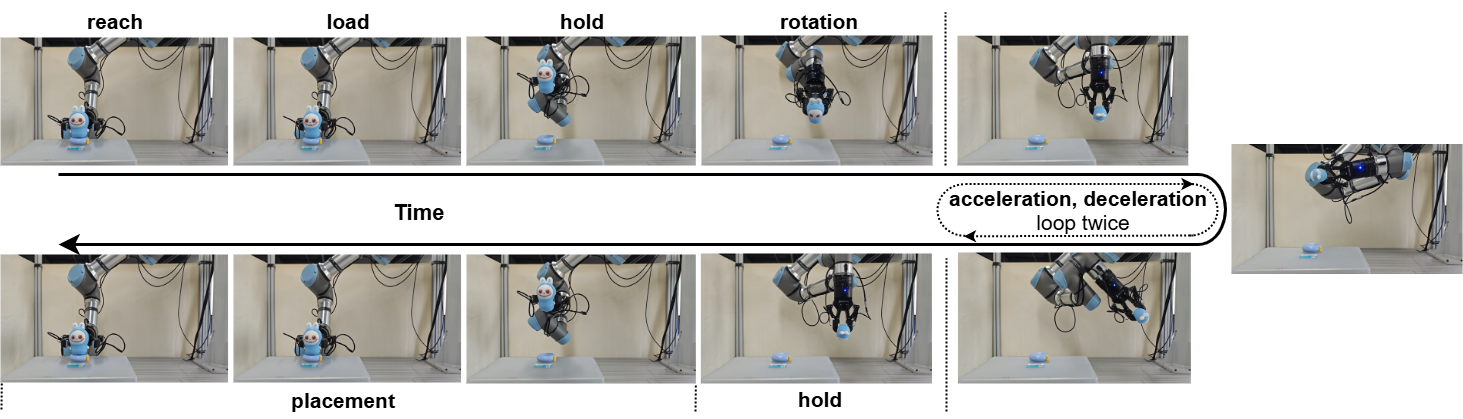}
    \caption{Grasp objects stably while robot moving. Robot picks up a soft toy which has high compliance. Then robot grasp stably while doing acceleration and deceleration movement. The end-effector reached a maximum speed of 3.0 $m/s$ and a maximum acceleration of 3.0 $m/s^{2}$.}
    \label{fig-grasp labobo shake}
\end{figure*}

\begin{figure*}
    \centering
    \includegraphics[width=\linewidth]{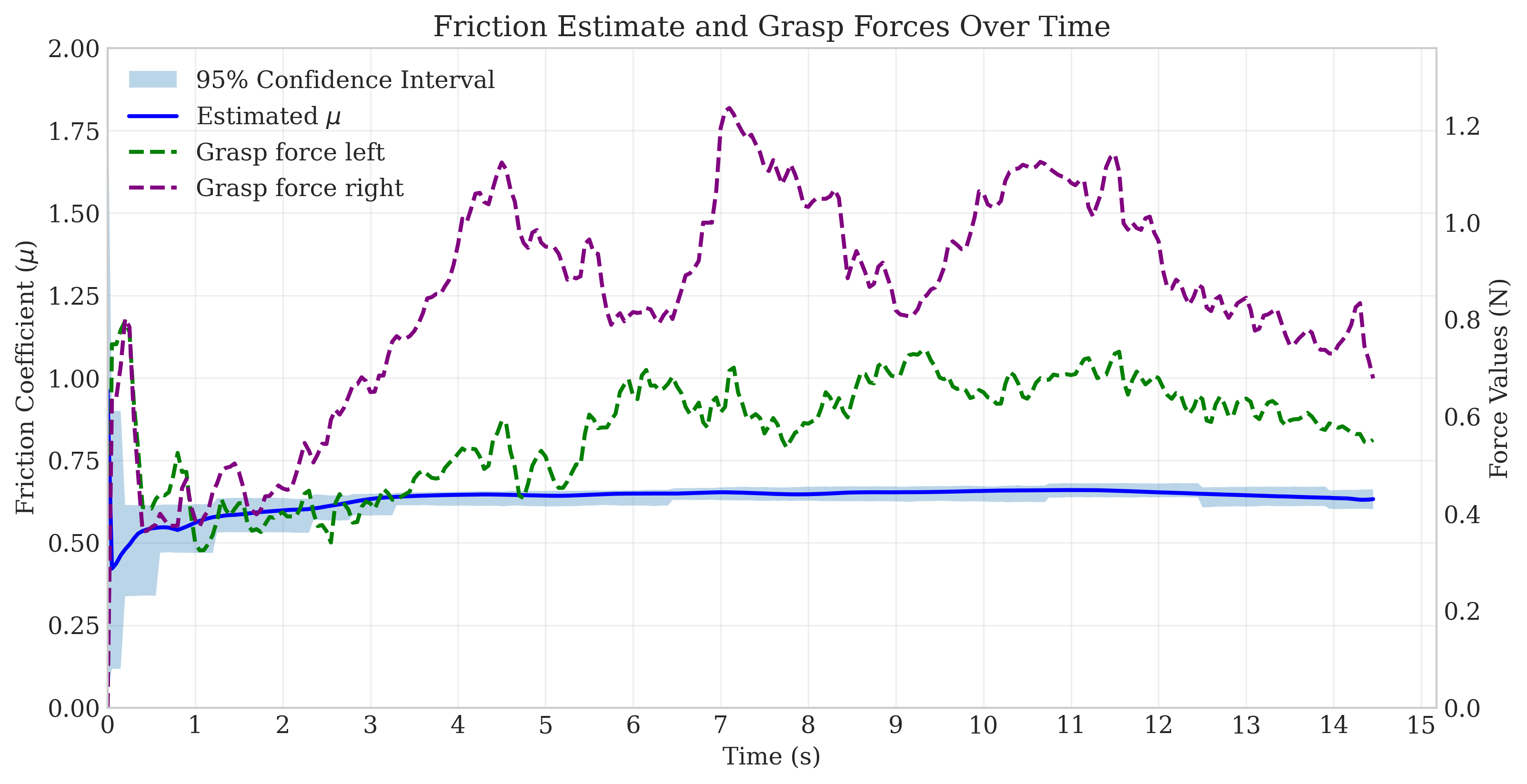}
    \caption{The recording of grasp forces and friction estimation during grasp while robot moving. The forces are adjusted based on dynamic movement conditions to keep stable grasp while also keeps small enough grasp forces.}
    \label{fig-soft labobo shake grasp}
\end{figure*}

\subsection{Stable Grasp for Objects with Unstable Friction and Changing Weight}

To evaluate the robustness of the proposed framework, a grasping experiment is conducted using a plastic cup with a moistened surface, while water is incrementally added during the manipulation task. This scenario induces dynamic variations in both surface friction and object mass, presenting a challenging condition even for human grasping.

The entire process is shown in Fig. \ref{fig-grasp plastic cup} and unfolds through the following sequential stages. During the (1) reach phase, the gripper closes until the measured contact force exceeds the threshold. Subsequently, in the (2) load phase, the synchronized online friction estimation and adaptive grasp control begin works. The (3) lift phase follows, during which the robot raises the object to a target position and (4) holds it steadily throughout the (5) watering stage. It is clearly observable that the grasp force is dynamically adjusted in response to the pouring of water in stages (5) and (6) to maintain a stable grip. Finally, the filled cup is transported and returned to its original position from stages (7) to (9).


The recording of estimation process and corresponding grasp force adjustments is illustrated in Fig. \ref{fig-grasp plastic cup}. The friction coefficient estimate converges rapidly as the grasping process progresses, stabilizing within a single time step after initial contact. This can be observed clearly in  supplementary videos. It is important to note that the force adjustment and friction estimation are tightly synchronized: at each time step, the current estimate directly informs the grip force modulation, while new tactile observations simultaneously refine the estimation. This closed-loop coordination enables efficient and stable grasping under dynamic conditions—a capability that mirrors human grasping behavior. 

As shown in the Fig. \ref{fig-results plastic cup grasp}, grasp stability is maintained even as water is poured into the cup (approximately from 5s to 8s). During this period, the grasping force increases adaptively, while the friction estimate exhibits minor variations—reflected in the confidence interval—which can be attributed to changes in surface moisture and total mass. Despite these perturbations, the system consistently maintains a stable grip until task completion.

\subsection{Grasp Objects Stably While Robot Moving}

Robots must not only learn to grasp objects effectively, but also maintain stable grasps during movement. In this section, we evaluate the proposed grasping framework under dynamic conditions wherein the robot executes acceleration and deceleration motions while holding various deformable objects.

The entire process is illustrated in Fig. \ref{fig-grasp labobo shake}. The robot picks up the object, performs repeated acceleration and deceleration movements, and returns the object to its initial position. The end-effector reached a maximum speed of 3.0 $m/s$ and a maximum acceleration of 3.0 $m/s^{2}$ during this experiment. As shown in Fig. \ref{fig-soft labobo shake grasp}, which records the grasp force and friction estimation during the process, the controller continuously adjusts grip forces in response to motion-induced disturbances to maintain grasp stability. Although minor variations in the friction estimate occur due to motion dynamics, the estimation remains stable throughout the task. The results presented here correspond to the soft toy object; more experimental results for other objects are available in the supplementary material.


\section{CONCLUSIONS}
\label{sec-conclusion}

In this paper, we presented a new VBTS-based framework that achieves gentle and stable manipulation of delicate objects through tight integration of real-time friction estimation and adaptive grasp control. The proposed system enables robust grasping performance by continuously synchronizing physical interaction feedback with closed-loop force modulation. Extensive robotic experiments validate the stability and effectiveness of our approach across a variety of delicate objects and highly dynamic conditions.

While the framework demonstrates reliable performance, it currently requires predefining the contact threshold force. Future work will focus on automating this parameter selection through foundation models \cite{xie_deligrasp_24}, potentially enabling the system to autonomously predict and update contact thresholds during operation and retain this knowledge for subsequent grasp attempts.




\addtolength{\textheight}{-12cm}   








\bibliographystyle{IEEEtran}
\bibliography{references}

\end{document}